\pdfoutput=1

\documentclass[11pt]{article}

\usepackage[]{acl}

\usepackage{times}
\usepackage{latexsym}

\usepackage[T1]{fontenc}
\usepackage[utf8]{inputenc}
\usepackage{microtype}
\usepackage{inconsolata}

\usepackage{graphicx}
\usepackage{longtable}
\usepackage{booktabs}
\usepackage{array}
\usepackage{enumitem}
\usepackage{amssymb}
\usepackage{wasysym}

\usepackage{url}

\title{Harnessing Large Language Models for Disaster Management: A Survey}

\author{
    Zhenyu Lei$^{\blacklozenge}$ \:
    Yushun Dong$^\Diamond$\thanks{Corresponding authors}\:  Weiyu Li$^\heartsuit$ \\ \bf
    Rong Ding$^\heartsuit$ \:
    Qi Wang$^\heartsuit$\footnotemark[1] \:
    Jundong Li$^\blacklozenge$\footnotemark[1] \\
    $^\blacklozenge$University of Virginia, $^\Diamond$Florida State University, $^\heartsuit$Northeastern University\\
    \texttt{\{vjd5zr, jundong\}@virginia.edu}, \texttt{yd24f@fsu.edu} \\ \texttt{\{weiy.li, ding.ro, q.wang\}@northeastern.edu
} \\
}

\begin{document}
\maketitle
\begin{abstract}
Large Language Models (LLMs) have demonstrated remarkable capabilities across various domains, including their emerging role in mitigating threats to human life, infrastructure, and the environment during natural disasters. Despite increasing research on disaster-focused LLMs, there remains a lack of systematic reviews and in-depth analyses of their applications in natural disaster management. To address this gap, this paper presents a comprehensive survey of LLMs in disaster response, introducing a taxonomy that categorizes existing works based on disaster phases and application scenarios. By compiling public datasets and identifying key challenges and opportunities, this study aims to provide valuable insights for the research community and practitioners in developing advanced LLM-driven solutions to enhance resilience against natural disasters.
\end{abstract}

\section{Introduction}
Natural disasters are becoming increasingly frequent and severe, posing unprecedented threats to human life, infrastructure, and the environment~\cite{manyena2006concept, yu2018big, chaudhary2021natural}. The 2010 Haiti earthquake, for instance, resulted in over 200,000 fatalities and widespread infrastructure devastation~\cite{desroches2011overview}. Similarly, the 2020 Australian bushfires caused the deaths of at least 33 people and an estimated loss of one billion animals~\cite{deb2020causes}. The profound impact of such catastrophic events underscores the urgent need for effective disaster management strategies.
Recently, large language models (LLMs) have transformed research and technological innovation with their exceptional capabilities in contextual understanding, logical reasoning, and complex problem-solving across multiple modalities~\cite{zhang2024comprehensive, zhang2024scientific, wei2025large}.
These capabilities position LLMs as powerful tools for natural disaster management, enabling them to analyze vast real-time disaster data, facilitate dynamic communication with affected communities, and support critical decision-making~\cite{otal2024llm}.

Despite their potential, a systematic review of LLMs in disaster management remains absent, limiting researchers and practitioners in identifying best practices, addressing research gaps, and optimizing LLM deployment for disaster-related challenges. To bridge this gap, this paper presents a comprehensive survey of LLM applications in disaster management, categorizing them across three model architectures and the four key disaster phases: mitigation, preparedness, response, and recovery. We introduce a novel taxonomy that integrates application scenarios, specific tasks, and model architectures tailored to disaster-related challenges. Additionally, we summarize publicly available datasets, identify key challenges, and explore avenues for enhancing the effectiveness, efficiency, and trustworthiness of LLMs in disaster response. This review aims to inspire and guide AI researchers, policymakers, and practitioners toward developing LLM-driven disaster management frameworks. Our key contributions are as follows:


\begin{itemize}
    \item \textbf{Systematical Review:} We provide the first systematical review of explorations of LLMs applications in disaster management across four key disaster phases.
    \item \textbf{Novel Taxonomy:} We propose a taxonomy integrating application scenarios, specific tasks, and model architectures, providing both practical and technical insights into this survey.
    \item \textbf{Resource Compilation:} We compile essential resources (e.g., datasets), and highlight key challenges and future research directions to advance LLM-driven disaster management.
\end{itemize}

\section{Background}

Disaster management is a multidisciplinary field that integrates resources, expertise, and strategies to mitigate the impact of increasingly severe disasters. Its primary goal is to minimize immediate damage while fostering long-term resilience and adaptive recovery.
Disaster management comprises four interconnected phases~\cite{sun2020applications}: 
\begin{itemize}
    \item \textbf{Mitigation} involves identifying risks and vulnerabilities while implementing proactive measures to prevent disasters.
    \item \textbf{Preparedness} includes developing comprehensive plans and public education initiatives to enhance readiness for potential disasters.
    \item \textbf{Response} identifies and addresses immediate needs during a disaster, including emergency rescue operations and resource distribution.
    \item \textbf{Recovery} involves rebuilding affected areas, addressing both physical and social impacts to facilitate a return to normalcy. 
\end{itemize}

In general, LLMs have the potential to serve as general-purpose foundations for developing specialized AI tools that enhance various aspects of disaster management.
%
Here, we categorize LLM architectures into three main types: 
(1) encoder-based LLM (e.g., BERT~\cite{devlin2018bert}), which excel in contextual understanding; (2) (encoder-)decoder LLM (e.g., GPT~\cite{brown2020language}), which are optimized for sequential prediction; and (3) multimodal LLMs, which integrate multiple modalities to enhance information processing~\cite{tiong2022plug, madichetty2021multi}
In disaster management, common downstream tasks include classification (e.g., damage classification), estimation (e.g., severity estimation), extraction (e.g., knowledge extraction), and generation (e.g., report generation). To tailor LLMs for these tasks, techniques such as fine-tuning and prompting are commonly employed.

\section{LLM For Disaster Management}
Foundation models can be utilized across the four disaster management phases: mitigation, preparedness, response, and recovery. Within each phase, existing works are categorized based on application scenarios, specific tasks, and model architectures. Figure~\ref{fig:overview} presents an overview of our taxonomy, with detailed summaries provided in Appendix~\ref{app:A}.

\subsection{Disaster Mitigation}
Assessing vulnerabilities is a crucial component of disaster mitigation, where LLMs have demonstrated promising potential. This process involves identifying and analyzing infrastructure and communities at risk, enabling proactive measures to reduce disaster impact.



\vspace{3pt}
\noindent\textbf{\textit{Vulnerability} Classification.} 
A system named \textit{Infrastructure Ombudsman} has leveraged supervised learning with encoder-based LLMs and zero-shot prompt learning with (encoder-)decoder LLMs to detect and classify concerns about potential infrastructure failures from social media data~\cite{chowdhury2024infrastructure}. This approach enables decision-makers to effectively prioritize resources and address critical issues in a timely manner.

\vspace{3pt}
\noindent\textbf{\textit{Answer} Generation.} Beyond infrastructure vulnerability assessment, (encoder-)decoder LLMs can help address community vulnerability-related queries by retrieving and leveraging the Social Vulnerability Index (SVI)~\cite{martelo2024towards}.

\subsection{Disaster Preparedness}
In the long term, LLMs can play a pivotal role in disaster preparedness through (1) enhancing public awareness by disseminating accurate and accessible information, and (2) supporting disaster forecasting with advanced data analysis. Building on these forecasts, LLMs can aid decision-makers in issuing (3) timely disaster warnings, improving short-term preparedness. Furthermore, LLMs can support well-structured (4) evacuation planning, ensuring the safe relocation of individuals and assets.

\begin{figure*}
    \centering
    \includegraphics[width=\linewidth]{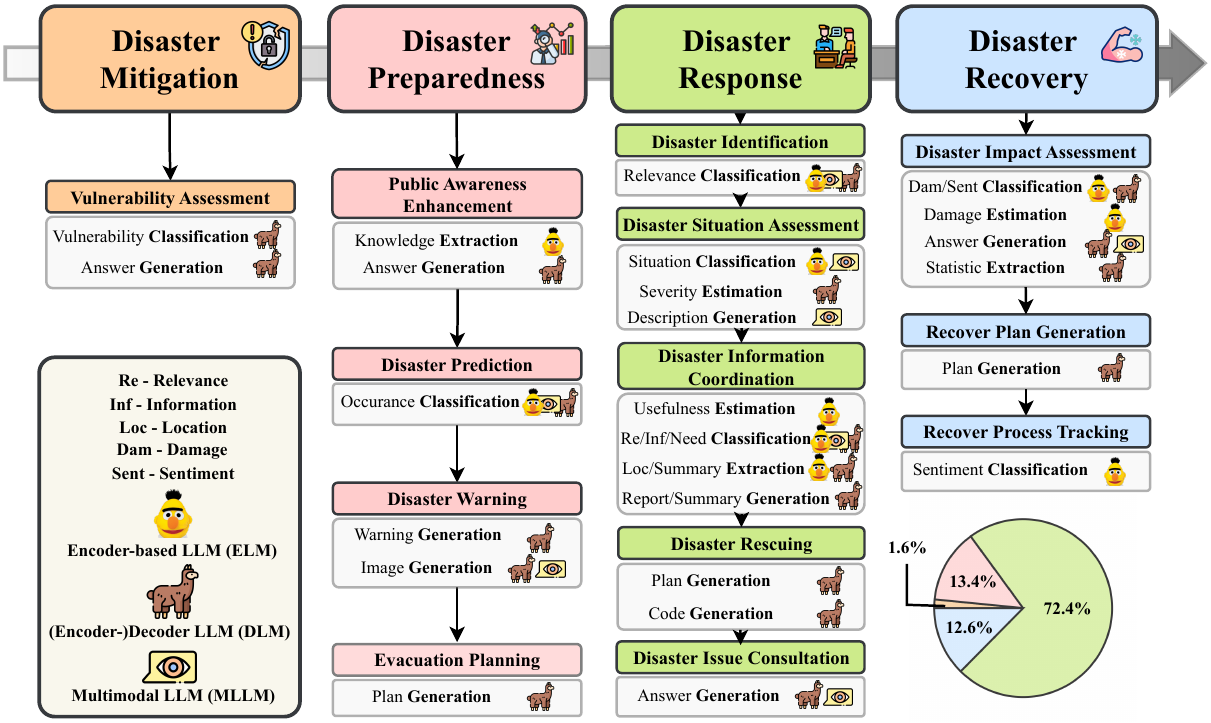}
    \caption{Taxonomy of applications of LLMs in disaster management. This survey categorizes the utilization of LLMs across four \textbf{disaster phases}, highlighting specific \textbf{applications} where \textbf{tasks} such as classification, estimation, extraction, and generation are performed by three \textbf{types of LLMs} (Encoder-based, (Encoder-)Decoder, and Multimodal LLM). The chart in the bottom-right corner presents the distribution of papers across each phase.}
    \label{fig:overview}
\end{figure*}

\subsubsection{Public Awareness Enhancement}
Enhancing public awareness of disasters is crucial, particularly by providing insights and knowledge derived from past disaster experiences.

\vspace{3pt}
\noindent\textbf{\textit{Knowledge} Extraction.}
\textit{Encoder-based LLMs} have been fine-tuned to extract disaster-related knowledge from news articles and social media~\cite{fu2024extracting}, as well as from extensive disaster literature~\cite{zhang2023automatic}, using Named Entity Recognition (NER). 
To improve the logical coherence of extracted entities, Ma et al. propose BERT-BiGRU-CRF for NER, enabling the construction of disaster knowledge graphs~\cite{ma2023ontology}. 
In addition, \textit{(encoder-)decoder LLMs} have been fine-tuned with instructional learning to extract knowledge triplets from documents for knowledge graph construction~\cite{wu2024intelligent}. 

\vspace{3pt}
\noindent\textbf{\textit{Answer} Generation.}
The extracted disaster knowledge could be incorporated in \textit{(encoder-)decoder LLMs}' prompts, facilitating disaster-related question answering~\cite{hostetter2024large, martelo2024towards, li2023ai}. Additionally, techniques such as retrieval-augmented generation (RAG) have been employed to further improve knowledge integration~\cite{zhu2024flood}.

\subsubsection{Disaster Prediction}
Effective disaster preparedness also relies on accurate and reliable disaster prediction.

\vspace{3pt}
\noindent\textbf{\textit{Occurrence} Classification.} \textit{Encoder-based LLMs} have been widely employed for disaster prediction. For instance, BERT has been integrated with GRU and CNN to predict disasters~\cite{indra2023modeling}. However, textual data alone is often limited due to its subjective and imprecise nature, prompting the adoption of \textit{multimodal LLMs} that incorporate multiple data modalities.
For instance, Zeng et al. combine historical flood data with geographical descriptions of specific locations to assess disaster risk~\cite{zeng2023global}.
Additionally, satellite imagery has been leveraged to provide visual context, enhancing predictive accuracy~\cite{liu2023harnessing}. 
To further improve disaster prediction with explicit external knowledge, \textit{(encoder-)decoder LLMs} have been integrated with RAG to retrieve historical flood data, aiding in risk assessment and action recommendation~\cite{wang2024remflow}. 


\subsubsection{Disaster Warning}
Once a disaster is anticipated, timely warnings are essential for ensuring public safety.

\vspace{3pt}
\noindent\textbf{\textit{Warning} Generation.}
\textit{(Encoder-)decoder LLMs} have proven valuable in generating warning messages based on rule-based alerts derived from streaming data~\cite{chandra2024decision}, significantly improving the responsiveness of warning systems. Additionally, RAG has enhanced LLMs by enabling the retrieval of disaster alerts from official APIs, providing real-time information on impending disasters~\cite{martelo2024towards}.

\vspace{3pt}
\noindent\textbf{\textit{Image} Generation.}
In addition to textual warnings, visual warnings can provide more vivid and intuitive descriptions, effectively reaching a broader audience.
To achieve this, \textit{multimodal LLMs} enhanced by diffusion-based text-to-image generative models can generate detailed visual representations of impending disasters~\cite{korea2024text}, enhancing the clarity and impact of disaster alerts.

\subsubsection{Evacuation Planning}
\paragraph{\textit{Plan} Generation.} To safeguard individuals and property from impending disasters, \textit{(encoder-)decoder LLMs} have been prompted to generate escape plans and provide evacuation recommendations~\cite{hostetter2024large}.

\subsection{Disaster Response}


With accurate and real-time (1) disaster identification and (2) situation assessment, decision-makers can acquire critical insights to establish a solid foundation for response efforts. Additionally, LLMs can facilitate (3) disaster information coordination, enhancing collaboration among stakeholders for more effective disaster response. As a result, decision-makers can leverage LLMs to execute key actions, including (4) disaster rescue operations and (5) disaster-related consultations.

\subsubsection{Disaster Identification}
\label{disaster_identification}
Effective disaster response begins with accurate and real-time identification, enabling efficient interventions~\cite{said2019natural, weber2020detecting}. Social media serves as a valuable resource in this process, offering real-time updates from affected individuals~\cite{anderson2016governing, trono2015dtn}. 
%

\noindent\textbf{\textit{Relevance} Classification with Encoder-based LLMs.} Classifying social media posts to identify disaster-related content is a crucial step in disaster detection, where LLMs have proven to be highly effective. Encoder-based LLMs augmented with trainable adapters are commonly employed for this task through fine-tuning on annotated disaster corpora~\cite{ningsih2021disaster, singh2022twitter, lamsal2024crisistransformers}. 
%
Recognizing the diverse sources of disaster data, ensemble methods combine predictions from multiple LLMs to leverage their complementary strengths in processing varied linguistic patterns~\cite{mukhtiar2023relevance}. Pure LLM-based approaches may struggle to capture fine-grained structural features in disaster-related posts. To address this, hybrid architectures integrate CNNs to capture local n-gram patterns~\cite{franceschini2024detecting, song2021sentiment, meghatria2024harnessing}, attention-based BiLSTMs to model sequential dependencies~\cite{huang2022early}, and graph neural networks (GNNs) to represent semantic word relationships~\cite{manthena2023leveraging, ghosh2022gnom}.
To tackle the challenge of limited labeled training data, active learning has been employed to automatically label informative samples~\cite{paul2023fine}.

\noindent\textbf{\textit{Relevance} Classification with Encoder-based LLMs.} 
Furthermore, (encoder-)decoder LLMs such as GPT-4 have demonstrated strong performance in relevance classification using prompt learning techniques~\cite{taghian2023disaster}.
%

\noindent\textbf{\textit{Relevance} Classification with Multimodal LLMs.}
Image data also provide valuable insights for disaster analysis and can be integrated to enhance classification using multimodal LLMs. This integration can be achieved through simple aggregation~\cite{kamoji2023fusion, madichetty2021multi} or attention-based mechanisms~\cite{shetty2024disaster}. To address challenges arising from multimodal heterogeneity, Zhou et al. employ a Cycle-GAN combined with a mixed fusion strategy~\cite{zhou2023public}.
%
Beyond multimodal heterogeneity, research also tackles other challenges in multimodal learning. These include addressing label scarcity through semi-supervised minimax entropy domain adaptation frameworks~\cite{wang2022bert} and enhancing model performance by leveraging the complementary strengths of diverse LLMs and visual models using ensemble methods~\cite{hanif2023deepsdc}.
Beyond social media, data from sources such as satellite imagery and news articles can further enhance disaster analysis~\cite{jang2024multimodal}.

\subsubsection{Disaster Situation Assessment}

After disaster identification, assessing its severity and spread is essential for formulating effective response strategies.

\vspace{3pt}
\noindent\textbf{\textit{Situation} Classification.}
\textit{encoder-based LLMs} have been fine-tuned to for binary classification to identify situational posts~\cite{madichetty2021neural}. Raj et al. employ BERT and NER to extract disaster-related locations, using location counts as an indicator of disaster severity~\cite{raj2023flood}. Additionally, multimodal LLMs integrate visual data to further enhance disaster situational assessment~\cite{kanth2022deep}.

\vspace{3pt}
\noindent\textbf{\textit{Severity} Estimation.}
While classification provides only a coarse understanding, severity estimation offers precise quantitative insights.
\textit{(encoder-)decoder LLMs} enhanced with chain-of-thought (CoT) reasoning have been used to estimate earthquake intensity, expressed as Modified Mercalli Intensity (MMI)~\cite{mousavi2024gemini}. In addition, \textit{multimodal LLMs} use rich image data for more accurate estimations. For example, FloodDepth-GPT employs prompt-based guidance with GPT-4 to estimate floodwater depth from flood images.

\vspace{3pt}
\noindent\textbf{\textit{Description} Generation.}
Beyond categorical and statistical descriptions, multimodal LLMs can generate more comprehensible textual situational reports from disaster images~\cite{hu2024flood, wolf2023camera}.

\subsubsection{Disaster Information Coordination}
Coordinating disaster-related information is crucial for ensuring an organized and collaborative response~\cite{comfort2004coordination, bharosa2010challenges}. Social media plays a pivotal role in this process, as individuals actively share posts containing warnings, urgent needs, and other critical information~\cite{lindsay2011social, imran2015processing}.

\vspace{3pt}
\noindent\textbf{\textit{Usefulness} Estimation.} To improve the accessibility of valuable information, \textit{encoder-base LLMs} are utilized to filter informative tweets by computing usefulness ratings~\cite{yamamoto2022methods}. However, this approach requires a predefined threshold to determine the relevance of a tweet.

\vspace{3pt}
\noindent\textbf{\textit{Relevance} Classification.}
Several studies fine-tune \textit{encoder-based LLMs} for binary relevance classification, as discussed in Section~\ref{disaster_identification}. Additionally, LLMs have been applied to multi-level relevance classification to further refine disaster-related information filtering~\cite{blomeier2024drowning}.

\vspace{3pt}
\noindent\textbf{\textit{Information} Classification.}
To facilitate information dissemination, several studies have fine-tuned \textit{encoder-based LLMs} to classify posts based on different information types, including actionable types such as "important for managers"~\cite{sharma2021categorizing}; humanitarian types such as "Injured people"~\cite{yuan2022smart}; and disaster-specific types~\cite{liu2021crisisbert}.
When fine-tuning data is limited, augmentation strategies such as manual hashtag annotation~\cite{boros2022adapting} and self-training with soft labeling~\cite{li2021combining} are employed to enhance classification performance~\cite{lei2025st}.

Pure LLM-based methods may have limitations, as discussed in Section~\ref{disaster_identification}. In contrast, hybrid architectures enhance performance by integrating CNNs and BiLSTMs to improve local pattern comprehension~\cite{zou2024multi} and employing Graph Attention Networks (GATs) to capture correlations between tweet embeddings and information types~\cite{zahera2021aid}. Additionally, FF-BERT leverages an ensemble of BERT and CNN to combine model strengths for improved classification~\cite{wilkho2024ff}.
Other studies enhance the application of LLMs in disaster information classification by extracting rationales—evidence that supports classification decisions~\cite{nguyen2022towards, nguyen2023learning}. RACLC~\cite{nguyen2022rationale} employs a two-stage framework, utilizing contrastive learning to refine rationale extraction and improve classification performance.


%

\textit{(Encoder-)decoder LLMs} have also been employed for disaster type and humanitarian classification through instruction tuning~\cite{otal2024ai, yin2024crisissense}, as well as zero-shot and few-shot prompting~\cite{dinani2024disaster}.


\textit{Multimodal LLMs} can integrate rich data from social media to enhance classification by leveraging multiple modalities~\cite{lei2022bic}. This integration can be achieved through simple feature aggregation~\cite{zhang2022multimodal, yu2024multimodal} or more advanced fusion techniques, such as cross-attention mechanisms~\cite{abavisani2020multimodal} and dual transformer architectures~\cite{zhou2023visual}.
%
Additionally, Basit et al. classify posts into humanitarian or structural categories only when the text and image classification outputs align; otherwise, the posts are uninformative~\cite{basit2023natural}.

\vspace{3pt}
\noindent\textbf{\textit{Need} Classification.}
Social media enables individuals to express urgent needs during disasters. \textit{Encoder-based LLMs} have been employed to detect disaster-related needs~\cite{yang2024detection, vitiugin2024multilingual} and rescue requests~\cite{toraman2023tweets}.
%
Responders also use social media to share available resources. Encoder-based LLMs have been employed to match needs with resources using cosine similarity-based retrieval methods, where both offer and request posts are embedded using XLM-RoBERTa~\cite{conneau2019unsupervised}, optimizing resource allocation.

\vspace{3pt}
\noindent\textbf{\textit{Location} Extraction.}
Additionally, various post-processing techniques enhance information dissemination, particularly through location extraction.
Several studies fine-tune encoder-based LLMs for location reference recognition (LRR), classifying tokens into categories such as "Inside Locations" (ILOC) and "Other Tokens" (O)~\cite{mehmood2024named, suwaileh2022disaster, koshy2024applying}. LRR can be further improved by integrating a conditional random field (CRF) model, which enhances the logical consistency of extracted locations~\cite{ma2022chinese, zhang2021extracting}. Furthermore, external knowledge corpora can support location extraction. For instance, Caillaut et al. use cosine similarity to match post entities with a knowledge base, ensuring the authenticity of extracted locations~\cite{caillaut2024entity}.

\textit{(Encoder-)decoder LLMs} are widely used for extracting location-relevant information through prompt learning~\cite{yu2024multimodal}. To enhance accuracy, external knowledge has been incorporated into prompts, including geo-knowledge~\cite{hu2023geo} and Object Character Recognition-based object descriptions~\cite{firmansyah2024improving}.

%

\vspace{3pt}
\noindent\textbf{\textit{Summary} Extraction.}
Furthermore, summarizing disaster-related posts provides a macro-level understanding during crises. Several studies focus on identifying critical and informative posts for summarization by integrating advanced techniques into \textit{encoder-based LLMs}, such as integer linear programming (ILP)~\cite{nguyen2022rationale, nguyen2022crisicsum} and Rapid Automatic Keyword Extraction (RAKE)~\cite{garg2024ikdsumm}.

\noindent\textbf{\textit{Summary} Generation.}
\textit{(Encoder-)decoder LLMs} extend summarization capabilities by generating summaries from retrieved text. For example, Vitiugin et al. rank key tweets using an LSTM model and apply a T5 model to generate summaries based on the top-ranked tweets~\cite{vitiugin2022cross}. Crisis2Sum performs query-focused summarization through a multi-step process, including query-informed document retrieval, reranking, fact extraction, clustering, fusion into event nuggets, and final selection for summarization~\cite{seeberger2024crisis2sum}. Additionally, agent-based approaches can enhance summary quality by leveraging multiple LLMs for document retrieval, reranking, and instruction-following summarization~\cite{seeberger2024multi}.

\vspace{3pt}
\noindent\textbf{\textit{Report} Generation.}
\textit{(Encoder-)decoder LLMs} have been employed for disaster report generation, utilizing techniques such as RAG to extract relevant web data~\cite{colverd2023floodbrain} and Chain-of-Thought reasoning to enhance the coherence and accuracy of generated reports. 


\subsubsection{Disaster Rescuing}

Grounded in a comprehensive understanding of the disaster situation, disaster rescue focuses on saving lives and protecting property through timely and coordinated actions.

\vspace{3pt}
\noindent\textbf{\textit{Plan} Generation.}
Effective rescue operations require well-structured rescue plans. \textit{(Encoder-)decoder LLMs} are prompted to generate actionable response plans, offering essential guidance for disaster response~\cite{goecks2023disasterresponsegpt}.

\vspace{3pt}
\noindent\textbf{\textit{Code} Generation.}
Once a plan is established, \textit{(encoder-)decoder LLMs} can support its execution by assisting organizations and rescue teams. For instance, they can facilitate robotic system guidance during rescue operations by translating verbal inputs into actionable operational commands using RAG~\cite{panagopoulos2024selective}. 


\subsubsection{Disaster Issue Consultation}

During disasters, affected individuals and organizations often seek reliable guidance. Disaster issue consultation provides advice, safety updates, and expert recommendations, helping them access resources, evaluate options, and make informed decisions~\cite{jiang2024applications}.

\vspace{3pt}
\noindent\textbf{\textit{Answer} Generation.}
\textit{(Encoder-)decoder LLMs} are employed to generate answers for frequently asked questions and provide disaster-related guidance~\cite{rawat2024disasterqa, chen2024optimizing}. To mitigate hallucination, RAG is integrated with verified disaster-related documents. For example, WildfireGPT retrieves wildfire-related literature and data to enhance prompts~\cite{xie2024wildfiregpt}. Chen et al. introduce a prompt chain to guide LLM reasoning over a disaster knowledge graph, incorporating structured knowledge~\cite{chen2024enhancing}. Unlike traditional RAG approaches without training, Xia et al. combine fine-tuning for implicit knowledge updates with RAG for explicit knowledge, further improving response quality~\cite{xia2024question}.

Additionally, \textit{multi-modal LLMs} can integrate textual and visual data to enhance disaster response. For example, several visual question answering (VQA) models, such as Plug-and-Play VQA~\cite{tiong2022plug}, have been prompted for zero-shot VQA in disaster scenarios~\cite{sun2023unleashing}. To handle complex user queries, ADI introduces sequential modular tools, incorporating vision-language models (VLMs), object detection models, and semantic segmentation models~\cite{liu2024rescueadi}. Furthermore, FloodLense combines ChatGPT with diffusion models to highlight disaster-affected areas in images, enhancing flood-related geographical question answering~\cite{kumbam2024floodlense}.

\subsection{Disaster Recovery} 

LLMs can play a crucial role in (1) disaster impact assessment, a vital step in the recovery process. By providing a comprehensive understanding of disaster impacts, LLMs can assist decision-makers in (2) generating recovery plans tailored to specific needs. Additionally, disaster responders have leveraged LLMs for (3) continuous recovery process tracking, ensuring effectiveness and progress throughout the recovery phase.

\subsubsection{Disaster Impact Assessment}
Accurately assessing the extent of damage across both physical and social dimensions is essential for prioritizing recovery efforts effectively. 

\vspace{3pt}
\noindent\textbf{\textit{Damage} Classification.}
From the physical dimension, \textit{encoder-based LLMs} have been employed to identify and categorize disaster-related damage (e.g., human/infrastructure damage~\cite{malik2024categorization}, water/power supply damage~\cite{chen2021social}) 
%
Additionally, Zou et al. propose a BERT-BiLSTM-Sit-CNN framework, improving textual understanding for damage-related post identification and damage-type classification~\cite{zou2024multi}. Beyond type classification, LLMs have been utilized to assess damage severity. For instance, Jeba et al. employ BERT to classify damage impact severity in social media posts and news articles~\cite{jeba2024facebook}.

\vspace{3pt}
\noindent\textbf{\textit{Damage} Estimation.}
Damage severity can be more effectively quantified through fine-grained estimation. Chen et al. compute damage severity scores by measuring the similarity between post tokens and predefined seed words' embeddings, both of which are derived from encoder-based LLMs~\cite{chen2021social}.

\vspace{3pt}
\noindent\textbf{\textit{Answer} Generation.} 
In addition, \textit{(encoder-)decoder LLMs} can answer specific assessment questions. Ziaullah et al. employ RAG-enhanced LLMs to retrieve operational status updates of critical infrastructure facilities from social media data~\cite{ziaullah2024monitoring}. \textit{Multimodal LLMs} further incorporate remote sensing data for enhanced assessment. Estevao et al. prompt GPT-4o to generate damage assessments based on building images~\cite{estevao2024effectiveness}. To improve modality alignment, SAM-VQA employs a supervised attention-based vision-language model (VLM) to integrate image and question features for visual question answering (VQA) tasks~\cite{sarkar2023sam}. Additionally, auxiliary tasks have been leveraged to enhance VQA performance. For instance, DATWEP dynamically balances the significance of segmentation and VQA tasks by adjusting class weights during training~\cite{alsan2023dynamic}.

\vspace{3pt}
\noindent\textbf{\textit{Statistic} Extraction.}
\textit{(Encoder-)decoder LLMs} have also used few-shot learning to extract fatality information from social media~\cite{hou2022near}, offering timely insights into human loss.

\vspace{3pt}
\noindent\textbf{\textit{Sentiment} Classification.} From the social dimension, disasters can influence public sentiment, where \textit{encoder-based LLMs}~\cite{han2024enhanced, berbere2023exploring} have been fine-tuned to classify social media posts into positive and negative emotions.
In addition, Li et al. employ \textit{(encoder-)decoder LLM} (e.g., GPT 3.5) to classify posts into five emotional types, such as "panic" and "sadness", using zero-shot prompting~\cite{li2025mining}. This approach helps responders better understand and address the emotional impact of disasters.



\subsubsection{Recovery Plan Generation}
Based on impact assessment, a recovery plan is formulated to rebuild infrastructure, restore services, and strengthen resilience~\cite{hallegatte2018building}. 

\vspace{3pt}
\noindent\textbf{\textit{Plan} Generation.}
\textit{(Encoder-)decoder LLMs} have been applied in certain recovery scenarios to generate recovery and reconstruction plans. For example, ChatGPT has been prompted to develop disaster recovery strategies for business restoration~\cite{white2024small, lakhera2024leveraging}. 


\begin{figure}
    \centering
    \includegraphics[width=\linewidth]{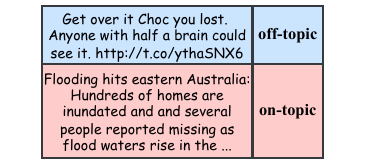}
    \caption{A sample of dataset for disaster relevance classification from CrisisLexT6~\cite{olteanu2014crisislex}.}
    \label{fig:dataset}
\end{figure}

\subsubsection{Recovery Process Tracking}
Continuous tracking of the recovery process ensures that progress remains aligned with the planned timeline, allowing decision-makers to adapt recovery strategies to evolving needs.

\vspace{3pt}
\noindent\textbf{\textit{Sentiment} Classification.}
\textit{Encoder-based LLMs} (e.g., BERTweet) have been employed to assess public sentiment throughout the recovery period~\cite{contrerasassessing}, enabling responders to tailor recovery efforts to effectively address the emotional needs of affected populations.

\begin{figure}[t]
    \centering
    \includegraphics[width=\linewidth]{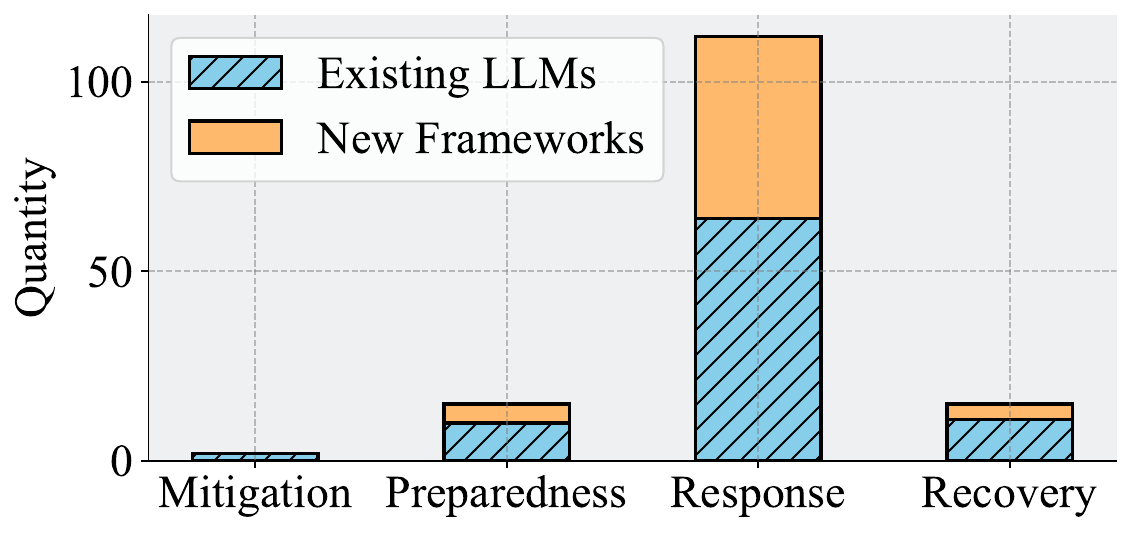}
    \caption{Publication counts utilizing existing LLMs and developing new models across the four phases of disaster management.}
    \vspace{-10pt}
    \label{fig:design}
\end{figure}

\section{Datasets}
Multiple disaster-related datasets have been employed to evaluate LLMs in disaster management. A comprehensive list of publicly available datasets is provided in Appendix~\ref{app:C}.

\vspace{3pt}
\noindent\textbf{Classification} datasets primarily consist of textual inputs from platforms such as Twitter and news outlets, categorizing data based on informativeness (relevance)~\cite{olteanu2014crisislex} (illustrated in Figure~\ref{fig:dataset}), humanitarian types~\cite{imran-etal-2016-twitter}, damage levels~\cite{alam2021crisisbench}, and other relevant attributes. Some datasets also incorporate visual data, including satellite imagery and social media images~\cite{alam2018crisismmd}. Model performance is typically evaluated using metrics such as accuracy and F1 score.


\vspace{3pt}
\noindent\textbf{Estimation} datasets usually provide quantitative labels such as flood depths~\cite{akinboyewa2024automated}. Metrics like Mean Absolute Error (MAE) are used for evaluation.

\vspace{3pt}
\noindent\textbf{Generation} datasets are also extensively used and primarily fall into two categories: question answering and summarization. Question-answering datasets provide disaster-related questions paired with crowdsourced annotated answers~\cite{rawat2024disasterqa}. Additionally, multimodal question-answering datasets, which incorporate disaster-related images as contextual information, are widely utilized~\cite{sun2023unleashing}. For the summarization task, large collections of documents serve as inputs, with reference summaries curated by domain experts~\cite{mccreadie2023crisisfacts}. Both question-answering and summarization tasks are evaluated using metrics such as BLEU.

\vspace{3pt}
\noindent\textbf{Extraction} datasets identify and label specific elements within a sentence, such as keywords~\cite{nguyen2022rationale} and locations~\cite{suwaileh2022disaster}. Tokens are labeled as "outside," "start," or "end" to indicate their extraction status. These datasets are primarily used for token-level classification tasks and are evaluated using classification metrics.

\section{Challenges and Opportunities}
Large Language Models (LLMs) hold great promise for disaster management but face several key limitations. Most studies deploy generic LLMs as universal solutions, overlooking domain-specific challenges and the need for tailored frameworks, as shown in Figure~\ref{fig:design}. Additionally, current applications are heavily concentrated on disaster response, leaving other phases underexplored, as illustrated in Figure~\ref{fig:overview}.
To fully harness the potential of LLMs in disaster management, researchers must address the disaster-specific challenges outlined below.



\vspace{3pt}
\noindent\textbf{Dataset Construction.} Current datasets are heavily skewed toward classification tasks, leaving other areas underexplored~\cite{proma2022nadbenchmarks}. Additionally, raw disaster data often contains uncertainty and bias~\cite{smith2013us}, posing challenges in constructing reliable datasets. Innovative approaches, such as synthetic data generation~\cite{kalluri2024robust}, offer a promising solution to enhance dataset coverage across diverse disaster scenarios.


\vspace{3pt}
\noindent\textbf{Efficient Deployment.} Large-scale LLMs face efficiency challenges~\cite{ramesh2024analyzing}, limiting their viability for real-time decision-making in emergency disaster scenarios. While lightweight models offer a more efficient alternative~\cite{saleem2024deltran15}, they often compromise robustness in disaster-related tasks. Developing models that balance efficiency and reliability is essential for effective disaster management.


\vspace{3pt}
\noindent\textbf{Robust Generation.} (Encoder-)decoder LLMs are prone to hallucination, generating factually inaccurate outputs that pose serious risks in disaster contexts, such as false evacuation routes, resource misallocation, and potential loss of lives. To mitigate these risks, strategies such as integrating RAG with external knowledge bases~\cite{colverd2023floodbrain}, domain-specific training~\cite{lamsal2024crisistransformers}, and uncertainty estimation~\cite{xu5037356enhancing} can help reduce hallucinated outputs and improve reliability.

\noindent\textbf{Unified Evaluation.} Most generative benchmarks for disaster (e.g. report/summary generation) rely on reference sets produced by domain experts. While these curated answers provide high-quality supervision, the underlying annotation criteria such as what counts as a correct answer often differ from one dataset to another. Consequently, published results are difficult to compare directly, because each study is implicitly tied to its own expert standard. As a result, it's important to build up a unified evaluation protocol to make more reliable comparisons.

\section{Conclusion}

This paper surveys the application of LLMs in disaster management across the four disaster phases, introducing a taxonomy that integrates application scenarios, specific tasks, and the architectures of models addressing these tasks. By presenting publicly available datasets and identifying key challenges, we aim to inspire collaborative efforts between AI researchers and decision-makers, ultimately enabling the full potential of LLMs to build more resilient communities and advance proactive disaster management practices.

\section*{Acknowledgment}
This work is supported in part by the National Science Foundation (NSF) under grants IIS-2006844, IIS-2144209, IIS-2223769, CNS-2154962, BCS-2228534, BCS-2228533, CMMI-2411248, CMMI-2125326, and CMMI-2402438. All authors would like to thank the reviewers and chairs for their constructive feedback and suggestions.

\section*{Limitations}
\noindent\textbf{Survey Scope.} This work focuses exclusively on disaster management applications only where existing LLMs have been utilized, leaving out other potential scenarios (e.g. repair cost evaluation during the recovery phase) that have yet to be explored in current research. In addition, we focus mostly on the natural disaster instead of man-made disaster. While these unexplored areas hold significant promise for future advancements, they fall beyond the scope of this study due to space constraints.

\noindent\textbf{Categorization.} This work categorizes papers based on their model architecture. However, it would be also beneficial to analyze existing papers from the other perspectives such as model size, inference efficiency, and model performance.

\noindent\textbf{Datasets.} Additionally, we include only a subset of datasets used in existing studies, prioritizing those that are easily accessible. Many datasets either are not open-sourced, have restrictive access policies, or lack assured quality, making them less suitable for reproducibility and further research.

\bibliography{acl_latex}

\appendix

\begin{figure*}
    \centering
    \includegraphics[width=\linewidth]{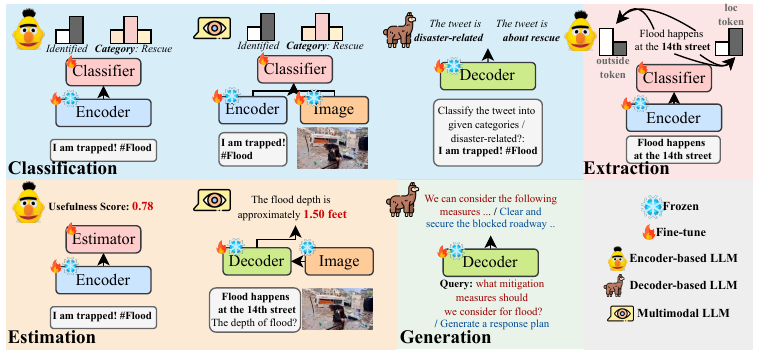}
    \caption{Pipeline of major tasks performed by different types of LLMs in disaster management.}
    \label{fig:pipeline}
\end{figure*}

\newpage
\section{Summary of Papers}
\label{app:A}
\subsection{Summary Table}
Table~\ref{tab:model} summarizes the surveyed papers, detailing their disaster phases, application scenarios, specific tasks, and architecture types.

\subsection{Pipeline Illustration}
\label{app:B}
In this section, we present Figure~\ref{fig:pipeline}, which illustrates the role of LLMs in disaster management. The figure outlines the major pipelines of three LLM architectures—encoder-based, (encoder-)decoder, and multimodal—applied across the four task types covered in this survey: classification, extraction, estimation, and generation. This visualization provides key insights into their mechanisms and applications in disaster management. 


\subsection{Statistics}
\label{app:D}
To provide a comprehensive overview of the current state of LLMs in disaster management, we present statistics from the surveyed papers, highlighting a significant gap between the NLP and disaster management communities. This gap underscores the urgent need for stronger interdisciplinary collaboration to bridge these fields and fully harness the potential of LLMs in addressing disaster-related challenges.

\begin{figure}[t]
    \centering
    \includegraphics[width=\linewidth]{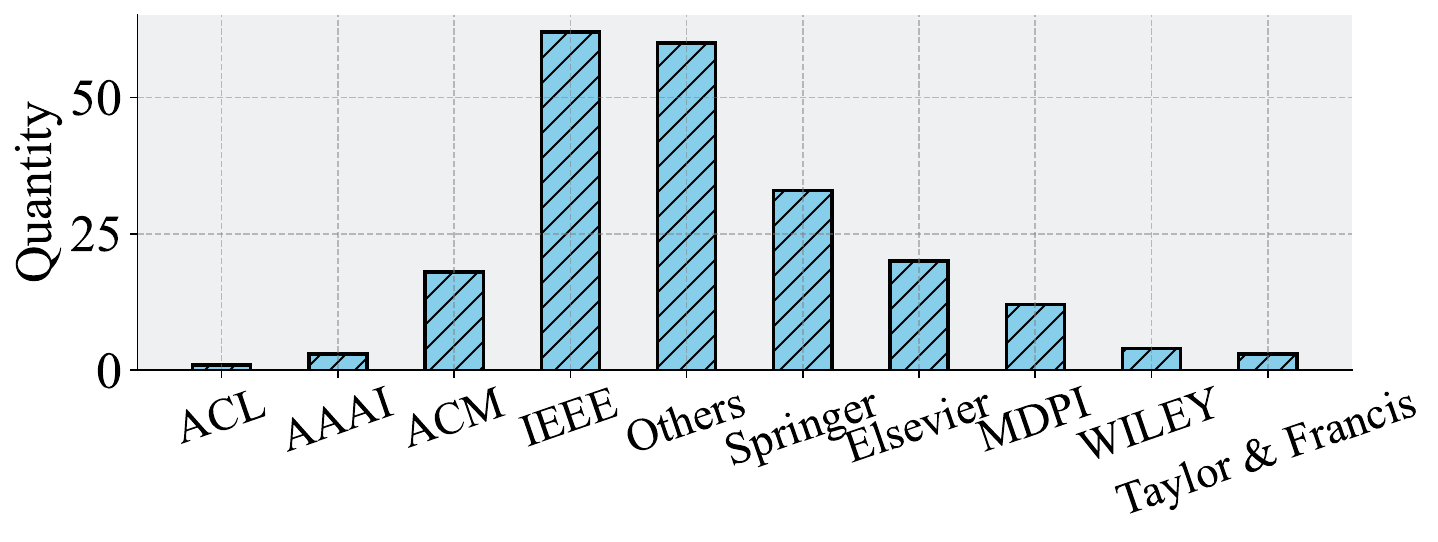}
    \caption{Publication distribution across various academic venues, with a focus on model design on the left and application-based research on the right.}
    \label{fig:quantity}
\end{figure}

Figure~\ref{fig:design} illustrates the number of publications leveraging existing LLMs versus those developing new frameworks, revealing that most studies are heavily application-focused. The majority rely on fine-tuning or prompting existing LLMs for disaster management tasks, rather than designing novel architectures. While some efforts have provided valuable insights, most research remains concentrated on the response phase, with limited exploration across other critical disaster management scenarios.
Figure~\ref{fig:quantity} illustrates the distribution of publications across academic venues, revealing that relatively few disaster management papers appear in NLP- or AI-specific conferences and journals. This trend reflects limited engagement from the LLM research community in this domain, underscoring the need to increase awareness and foster greater collaboration within the field. 

\section{Datasets}
\label{app:C}
Table~\ref{tab:dataset} summarizes existing publicly available datasets. For classification tasks, we exclude datasets that focus on a single disaster type if they are already incorporated into comprehensive benchmarks such as CrisisBench~\cite{alam2021crisisbench}.

\subsection{Classification Datasets}
\begin{itemize}
    \item \textbf{CrisisLexT6}~\cite{olteanu2014crisislex}: This dataset is designed for relevance classification. It contains data from six crisis events between October 2012 and July 2013.
    \item \textbf{CrisisLexT26}~\cite{olteanu2015expect}: This dataset is an updated version of CrisisLexT6, which contains public data from 26 crisis events in 2012 and 2013 with relevance information and six humanitarian categories.
    \item \textbf{CrisisNLP}~\cite{imran-etal-2016-twitter}: This dataset is a large-scale dataset that includes classes from humanitarian disaster responses and classes related to health emergencies. It is collected from 19 different disaster events that happened between 2013 and 2015.
    \item \textbf{SWDM2013}~\cite{imran2013practical}: This dataset is utilized for relevance classification that consists of tweets from two events: (i) the Joplin collection contains tweets from the tornado that struck Joplin, Missouri on May 22, 2011; (ii) The Sandy collection contains tweets collected from Hurricane Sandy that struck the Northeastern US on Oct 29, 2012. 
    \item \textbf{ISCRAM2013}~\cite{imran2013practical}: This dataset consists of tweets collected from the same events as in SWDM2013, containing both relevance and humanitarian categories.
    \item \textbf{Disaster Response Data (DRD)}~\cite{alam2021crisisbench}: This dataset consists of tweets collected during various crisis events that took place in 2010 and 2012. This dataset is annotated using 36 classes that include relevance as well as humanitarian categories.
    \item \textbf{Disasters on Social Media (DSM)}~\cite{alam2021crisisbench}: This dataset comprises 10K tweets annotated with relevance labels.
    \item \textbf{AIDR}~\cite{imran2014aidr}: This dataset contains data obtained from the AIDR system on September 25, 2013, collecting tweets using hashtags such as "\#earthquake". It is utilized for relevance and humanitarian classification.
    \item \textbf{CrisisMMD}~\cite{alam2018crisismmd}: This dataset is a multimodal and multitask dataset comprising $16k$ labeled tweets and corresponding images. Tweets have been sourced from seven natural disaster events that took place in 2017. Each sample is annotated with relevance, humanitarian (eight classes), and damage severity categories (mild, severe, and none).
    \item \textbf{Multi-Crisis}~\cite{sanchez2023cross}: This dataset was proposed to evaluate transfer learning scenarios where data from high-resource languages (e.g., English) is used to classify messages in low-resource languages (e.g., Spanish, Italian) and unseen crisis domains, with relevance and humanitarian categories. It is collected from 7 existing datasets, 53 crisis events, and contains 9 domains.
    \item \textbf{CrisisBench}~\cite{alam2021crisisbench}: This dataset is a comprehensive benchmark consolidated from 9 existing datasets, utilized for relevance and humanitarian classification.
    \item \textbf{Eyewitness Messages}~\cite{zahra2020automatic}: This dataset is designed to identify disaster eyewitness-related tweets and classify them into three categories: direct eyewitnesses, indirect eyewitnesses, and vulnerable eyewitnesses—individuals who anticipate a disaster and are present in regions where disaster warnings have been issued. It comprises 14,000 tweets collected from earthquakes, hurricanes, and wildfires.
    \item \textbf{TREC Incident Streams}~\cite{mccreadie2019trec}: This dataset has been developed as part of the TREC-IS 2018 evaluation challenge and consists
    of $20k$ tweets labeled for actionable information identification and information criticality assessment.
    \item \textbf{HumAID}~\cite{alam2021humaid}: This dataset contains $77k$ labeled tweets, which are sampled from 24 million tweets collected during 19 disasters between 2016 and 2019, including hurricanes, earthquakes, wildfires, and floods. It is balanced in terms of disaster types and contains 7 humanitarian categories.
    \item \textbf{EPIC}: This dataset contains data primarily collected from Hurricane Sandy, including tweets from 93 users across four annotation schemes, with data spanning three weeks around the hurricane’s landfall. It is used for relevance and humanitarian classification.
\end{itemize}

\subsection{Estimation Datasets}
\begin{itemize}
    \item \textbf{Did You Feel It (DYFI)}~\cite{atkinson2007did}: This dataset includes ground shaking intensity and geographic distribution information, collected from post-earthquake reports through 750,000 online questionnaire responses from individuals who experienced the event.
    \item \textbf{FloodDepth}~\cite{akinboyewa2024automated}: This dataset consists of 150 flood photos collected online, used to estimate floodwater depth based on various reference objects, including stop signs, vehicles, and humans.
    \item \textbf{Behavioral Facilitation (BF)}~\cite{yamamoto2022methods}: This dataset, collected after the 2018 Hokkaido earthquake, includes data labeled with usefulness ratings based on behavioral facilitation information.
\end{itemize}

\subsection{Extraction Datasets}
\begin{itemize}
    \item \textbf{\cite{fu2024extracting}}: This dataset contains county-level data from news media collected during urban flood events from 2000 to 2022. It is utilized to extract information such as the time and location of disasters.
    \item \textbf{\cite{ma2023ontology}}: This dataset is designed for entity and relation extraction, comprising 5,560 annotated instances, 12,980 entities, and 6,895 relations derived from reports on geological hazards.
    \item \textbf{DisasterMM}~\cite{andreadis2022disastermm}: This dataset was collected from Twitter by searching for flood-related keywords. It consists of two subsets: RCTP, which includes 6,672 tweets for relevance classification, and LETT, which contains 4,992 tweets used for location extraction. In the LETT subset, words are annotated with "B-LOC" for the first word in a sequence referring to a location, "I-LOC" for subsequent words within the same location sequence, and "O" for words that do not correspond to a location.
    \item \textbf{\cite{suwaileh2022disaster}}: This dataset contains 22,000 crisis-related tweets from various disasters, including floods, earthquakes, and hurricanes. It is annotated with location-related tags such as "inLOC" and "outLOC."
    \item \textbf{Re´SoCIO}~\cite{caillaut2024entity}: This dataset is constructed by merging Wikipedia datasets and multiple disaster-related datasets, annotated with a set of 9 NER labels with different types of information.
    \item \textbf{\cite{nguyen2022rationale}}: This dataset contains tweet data with annotated rationales from 4 subsets of CrisisNLP. It is used for rationale extraction, and the extracted rationales can assist in disaster classification.
\end{itemize}

\subsection{Generation Datasets}
\begin{itemize}
    \item \textbf{\cite{vitiugin2022cross}}: This dataset is used to generate summaries of various disaster events, with the official report of each event serving as the ground truth.
    \item \textbf{CrisisFACTS}~\cite{mccreadie2023crisisfacts}: This dataset is a multi-stream collection comprising data from eight crisis events gathered across various platforms. It is designed to process daily multi-platform streams and generate summaries based on specific information needs, such as "Have airports closed?"
    \item \textbf{DisasterQA}~\cite{rawat2024disasterqa}: This dataset includes disaster-related multiple choice questions from 7 different sources, examples could be "What causes a tsunami?".
    \item \textbf{FFD-IQA}~\cite{sun2023unleashing}: This dataset comprises 2,058 images and 22,422 question-meta ground truth pairs related to the safety of individuals trapped in disaster sites and the availability of emergency services. It includes three types of questions: free-form, multiple-choice, and yes-no questions.
    \item \textbf{FloodNet}~\cite{rahnemoonfar2021floodnet}: This dataset consists of 4,500 question-image pairs collected after Hurricane Harvey. The questions pertain to buildings, roads, and entire scenes, categorized into four groups: "Simple Counting," "Complex Counting," "Yes/No," and "Condition Recognition."
\end{itemize}





\clearpage
\onecolumn

\footnotesize
\begin{longtable}{@{}>{\raggedright\arraybackslash}p{3cm}p{1.5cm}p{4.3cm}p{3.4cm}p{1.2cm}p{0.6cm}@{}}
\caption{Summary of LLMs in disaster management with their disaster phases, application scenarios, specific tasks, and architecture types. "Arch": Type of LLM architectures used; "NM": Whether the paper presents novel methods.} \\ 
\vspace{-10mm}
\label{tab:model} \\
\toprule
\textbf{Paper} & \textbf{Phase} & \textbf{Application} & \textbf{Task} & \textbf{Arch} & \textbf{NM} \\ \midrule
\endfirsthead
\toprule
\textbf{Paper} & \textbf{Phase} & \textbf{Application} & \textbf{Task} & \textbf{Arch} & \textbf{NM} \\ \midrule
\endhead
\bottomrule
\endfoot
\cite{chowdhury2024infrastructure} & Mitigation & Vulnerability Assessment & Vulnerability Classification & Decoder & No\\
\cite{martelo2024towards} & Mitigation & Vulnerability Assessment & Answer Generation & Decoder & Yes\\
\cite{fu2024extracting} & Preparedness & Public Awareness Enhancement & Knowledge Extraction & Encoder & No \\
\cite{zhang2023automatic} & Preparedness & Public Awareness Enhancement & Knowledge Extraction & Encoder & No\\
\cite{ma2023ontology}  & Preparedness & Public Awareness Enhancement & Knowledge Extraction & Encoder & Yes \\
\cite{wu2024intelligent}  & Preparedness & Public Awareness Enhancement & Knowledge Extraction & Decoder & No\\
\cite{hostetter2024large}  & Preparedness & Public Awareness Enhancement & Answer Generation & Decoder& No \\
\cite{martelo2024towards} & Preparedness & Public Awareness Enhancement & Answer Generation & Decoder & No\\
\cite{li2023ai}  & Preparedness & Public Awareness Enhancement & Answer Generation & Decoder& No \\
\cite{indra2023modeling} & Preparedness & Disaster Forecast & Occurrence Classification & Encoder & Yes \\
\cite{zeng2023global} & Preparedness & Disaster Forecast & Occurrence Classification & Multimodal & Yes\\
\cite{liu2023harnessing} & Preparedness & Disaster Forecast & Occurrence Classification & Multimodal & Yes\\
\cite{wang2024remflow} & Preparedness & Disaster Forecast & Occurrence Classification & Decoder& No \\
\cite{chandra2024decision} & Preparedness & Disaster Warning & Warning Generation & Decoder & No\\
\cite{martelo2024towards} & Preparedness & Disaster Warning & Warning Generation & Decoder & No\\
\cite{korea2024text} & Preparedness & Disaster Warning & Image Generation & Multimodal & Yes\\
\cite{hostetter2024large} & Preparedness & Evacuation Planning & Plan Generation & Decoder & No\\
\cite{ningsih2021disaster} & Response & Disaster Identification & Relevance Classification & Encoder & No\\
\cite{madichetty2023roberta} & Response & Disaster Identification & Relevance Classification & Encoder & No\\
\cite{singh2022twitter} & Response & Disaster Identification & Relevance Classification & Encoder & No\\

\cite{powers2023using} & Response & Disaster Identification & Relevance Classification & Encoder & No\\
\cite{duraisamy2024twitter} & Response & Disaster Identification & Relevance Classification & Encoder & No\\
\cite{ullah2023unveiling} & Response & Disaster Identification & Relevance Classification & Encoder & No\\
\cite{li2024typhoon} & Response & Disaster Identification & Relevance Classification & Encoder& No \\
\cite{zhao2024information} & Response & Disaster Identification & Relevance Classification & Encoder& No \\
\cite{karanjit2024converging} & Response & Disaster Identification & Relevance Classification & Encoder & No\\
\cite{pabari2023flood} & Response & Disaster Identification & Relevance Classification & Encoder & No\\
\cite{de2019global} & Response & Disaster Identification & Relevance Classification & Encoder & No\\
\cite{zhao2024information} & Response & Disaster Identification & Relevance Classification & Encoder & No\\
\cite{wang2021transformer} & Response & Disaster Identification & Relevance Classification & Encoder & No\\
\cite{habib2024relevance} & Response & Disaster Identification & Relevance Classification & Encoder & No\\
\cite{liu2021crisisbert} & Response & Disaster Identification & Relevance Classification & Encoder & No\\
\cite{fontalis2023comparative} & Response & Disaster Identification & Relevance Classification & Encoder & No\\
\cite{mehmood2024named} & Response & Disaster Identification & Relevance Classification & Encoder & No\\
\cite{paul2023fine} & Response & Disaster Identification & Relevance Classification & Encoder & Yes\\
\cite{lamsal2024crisistransformers} & Response & Disaster Identification & Relevance Classification & Encoder & Yes\\
\cite{manthena2023leveraging} & Response & Disaster Identification & Relevance Classification & Encoder & Yes\\
\cite{danday2022twitter} & Response & Disaster Identification & Relevance Classification & Encoder & Yes\\
\cite{ghosh2022gnom} & Response & Disaster Identification & Relevance Classification & Encoder & Yes\\
\cite{taghian2023disaster} & Response & Disaster Identification & Relevance Classification & Decoder & No\\
\cite{kamoji2023fusion} & Response & Disaster Identification & Relevance Classification & Multimodal & Yes\\
\cite{madichetty2021multi} & Response & Disaster Identification & Relevance Classification & Multimodal & Yes\\
\cite{koshy2023multimodal} & Response & Disaster Identification & Relevance Classification & Multimodal & Yes\\
\cite{shetty2024disaster} & Response & Disaster Identification & Relevance Classification & Multimodal & Yes\\
\cite{zhou2023public} & Response & Disaster Identification & Relevance Classification & Multimodal & Yes\\
\cite{yu2024multimodal} & Response & Disaster Identification & Relevance Classification & Multimodal & Yes\\
\cite{zhang2022multimodal} & Response & Disaster Identification & Relevance Classification & Multimodal & Yes\\
\cite{kota2022multimodal} & Response & Disaster Identification & Relevance Classification & Multimodal & Yes\\
\cite{wang2022bert} & Response & Disaster Identification & Relevance Classification & Multimodal & Yes\\
\cite{hanif2023deepsdc} & Response & Disaster Identification & Relevance Classification & Multimodal & Yes\\
\cite{jang2024multimodal} & Response & Disaster Identification & Relevance Classification & Multimodal & Yes\\
\cite{madichetty2021neural} & Response & Disaster Situation Assessment & Situation Classification & Encoder & Yes\\
\cite{raj2023flood} & Response & Disaster Situation Assessment & Situation Classification & Encoder & Yes\\
\cite{kanth2022deep} & Response & Disaster Situation Assessment & Situation Classification & Multimodal & Yes\\
\cite{mousavi2024gemini} & Response & Disaster Situation Assessment & Severity Estimation & Decoder & No\\
\cite{akinboyewa2024automated} & Response & Disaster Situation Assessment & Severity Estimation & Multimodal & No\\
\cite{hu2024flood} & Response & Disaster Situation Assessment & Description Generation & Multimodal & No\\
\cite{wolf2023camera} & Response & Disaster Situation Assessment & Description Generation & Multimodal & No\\
\cite{yamamoto2022methods} & Response & Disaster Information Coordination & Usefulness Estimation & Encoder & No\\
\cite{blomeier2024drowning} & Response & Disaster Information Coordination & Relevance Classification & Encoder & No\\
\cite{adesokan2023tweetace} & Response & Disaster Information Coordination & Information Classification & Encoder & No\\
\cite{wahid2022topic2labels} & Response & Disaster Information Coordination & Information Classification & Encoder & No\\
\cite{chandrakala2022identifying} & Response & Disaster Information Coordination & Information Classification & Encoder & No\\
\cite{naaz2021sequence} & Response & Disaster Information Coordination & Information Classification & Encoder & No\\
\cite{du2023applicability} & Response & Disaster Information Coordination & Information Classification & Encoder & No\\
\cite{adesokan2023tweetace} & Response & Disaster Information Coordination & Information Classification & Encoder & No\\
\cite{han2024quakebert} & Response & Disaster Information Coordination & Information Classification & Encoder & No\\
\cite{sharma2021categorizing} & Response & Disaster Information Coordination & Information Classification & Encoder & No\\
\cite{yuan2022smart} & Response & Disaster Information Coordination & Information Classification & Encoder & No\\
\cite{liu2021crisisbert} & Response & Disaster Information Coordination & Information Classification & Encoder & No\\
\cite{boros2022adapting} & Response & Disaster Information Coordination & Information Classification & Encoder & Yes\\
\cite{li2021combining} & Response & Disaster Information Coordination & Information Classification & Encoder & Yes\\
\cite{zou2024multi} & Response & Disaster Information Coordination & Information Classification & Encoder & Yes\\
\cite{zahera2021aid} & Response & Disaster Information Coordination & Information Classification & Encoder & Yes\\
\cite{wilkho2024ff} & Response & Disaster Information Coordination & Information Classification & Encoder & Yes\\
\cite{nguyen2022towards} & Response & Disaster Information Coordination & Information Classification & Encoder & Yes\\
\cite{zou2023decrisismb} & Response & Disaster Information Coordination & Information Classification & Encoder & Yes\\
\cite{nguyen2023learning} & Response & Disaster Information Coordination & Information Classification & Encoder & Yes\\
\cite{nguyen2022rationale} & Response & Disaster Information Coordination & Information Classification & Encoder & Yes\\
\cite{dar2025social} & Response & Disaster Information Coordination & Information Classification & Encoder & Yes\\
\cite{otal2024ai} & Response & Disaster Information Coordination & Information Classification & Decoder & No\\
\cite{yin2024crisissense} & Response & Disaster Information Coordination & Information Classification & Decoder & No\\
\cite{dinani2024disaster} & Response & Disaster Information Coordination & Information Classification & Decoder & No\\
\cite{zhang2022multimodal} & Response & Disaster Information Coordination & Information Classification & Multimodal & Yes\\
\cite{yu2024multimodal} & Response & Disaster Information Coordination & Information Classification & Multimodal & Yes\\
\cite{shetty2024disaster} & Response & Disaster Information Coordination & Information Classification & Multimodal & Yes\\
\cite{abavisani2020multimodal} & Response & Disaster Information Coordination & Information Classification & Multimodal & Yes\\
\cite{zhou2023visual} & Response & Disaster Information Coordination & Information Classification & Multimodal & Yes\\
\cite{basit2023natural} & Response & Disaster Information Coordination & Information Classification & Multimodal & Yes\\
\cite{yang2024detection} & Response & Disaster Information Coordination & Need Classification & Encoder & No\\
\cite{toraman2023tweets} & Response & Disaster Information Coordination & Need Classification & Encoder & No\\
\cite{zhou2022victimfinder} & Response & Disaster Information Coordination & Need Classification & Encoder & No\\
\cite{vitiugin2024multilingual} & Response & Disaster Information Coordination & Need Classification & Encoder & Yes\\
\cite{conneau2019unsupervised} & Response & Disaster Information Coordination & Need Classification & Encoder & Yes\\
\cite{lamsal2024crema} & Response & Disaster Information Coordination & Need Classification & Encoder & Yes\\
\cite{mehmood2024named} & Response & Disaster Information Coordination & Location Extraction & Encoder & No\\
\cite{suwaileh2022disaster} & Response & Disaster Information Coordination & Location Extraction & Encoder & No\\
\cite{koshy2024applying} & Response & Disaster Information Coordination & Location Extraction & Encoder & Yes\\
\cite{ma2022chinese} & Response & Disaster Information Coordination & Location Extraction & Encoder & Yes\\
\cite{zhang2021extracting} & Response & Disaster Information Coordination & Location Extraction & Encoder & Yes\\
\cite{caillaut2024entity} & Response & Disaster Information Coordination & Location Extraction & Encoder & Yes\\
\cite{yu2024multimodal} & Response & Disaster Information Coordination & Location Extraction & Decoder & No\\
\cite{hu2023geo} & Response & Disaster Information Coordination & Location Extraction & Decoder & No\\
\cite{firmansyah2024improving} & Response & Disaster Information Coordination & Location Extraction & Decoder & No\\
\cite{nguyen2022rationale} & Response & Disaster Information Coordination & Summary Extraction & Encoder & Yes\\
\cite{nguyen2022crisicsum} & Response & Disaster Information Coordination & Summary Extraction & Encoder & Yes\\
\cite{garg2024ikdsumm} & Response & Disaster Information Coordination & Summary Extraction & Encoder & Yes\\
\cite{vitiugin2022cross} & Response & Disaster Information Coordination & Summary Extraction & Decoder & Yes\\
\cite{colverd2023floodbrain} & Response & Disaster Information Coordination & Report Generation & Decoder & No\\
\cite{pereira2023crisis} & Response & Disaster Information Coordination & Report Generation & Decoder & No\\
\cite{seeberger2024multi} & Response & Disaster Information Coordination & Report Generation & Decoder & Yes\\
\cite{seeberger2024crisis2sum} & Response & Disaster Information Coordination & Report Generation & Decoder & Yes\\
\cite{goecks2023disasterresponsegpt} & Response & Disaster Rescuing & Plan Generation & Decoder & No\\
\cite{panagopoulos2024selective} & Response & Disaster Rescuing & Code Generation & Decoder & No\\
\cite{rawat2024disasterqa} & Response & Disaster Issue Consultation & Answer Generation & Decoder & No\\
\cite{chen2024optimizing} & Response & Disaster Issue Consultation & Answer Generation & Decoder & No\\
\cite{xie2024wildfiregpt} & Response & Disaster Issue Consultation & Answer Generation & Decoder & No\\
\cite{chen2024enhancing} & Response & Disaster Issue Consultation & Answer Generation & Decoder & Yes\\
\cite{xia2024question} & Response & Disaster Issue Consultation & Answer Generation & Decoder & Yes\\
\cite{sun2023unleashing} & Response & Disaster Issue Consultation & Answer Generation & Multimodal & No\\
\cite{liu2024rescueadi} & Response & Disaster Issue Consultation & Answer Generation & Multimodal & Yes\\
\cite{kumbam2024floodlense} & Response & Disaster Issue Consultation & Answer Generation & Multimodal & Yes\\
\cite{malik2024categorization} & Recovery & Disaster Impact Assessment & Damage Classification & Encoder & No\\
\cite{chen2021social} & Recovery & Disaster Impact Assessment & Damage Classification & Encoder & No\\
\cite{jeba2024facebook} & Recovery & Disaster Impact Assessment & Damage Classification & Encoder & No\\
\cite{zou2024multi} & Recovery & Disaster Impact Assessment & Damage Classification & Encoder & Yes\\
\cite{chen2021social} & Recovery & Disaster Impact Assessment & Damage Estimation & Encoder & Yes\\
\cite{ziaullah2024monitoring} & Recovery & Disaster Impact Assessment & Answer Generation & Decoder & No\\
\cite{estevao2024effectiveness} & Recovery & Disaster Impact Assessment & Answer Generation & Multimodal & No\\
\cite{sarkar2023sam} & Recovery & Disaster Impact Assessment & Answer Generation & Multimodal & No\\
\cite{alsan2023dynamic} & Recovery & Disaster Impact Assessment & Answer Generation & Multimodal & No\\
\cite{hou2022near} & Recovery & Disaster Impact Assessment & Statistic Extraction & Decoder & No\\
\cite{han2024enhanced} & Recovery & Disaster Impact Assessment & Sentiment Classification & Encoder & No\\
\cite{alharm4755638enhancing} & Recovery & Disaster Impact Assessment & Sentiment Classification & Encoder & No\\
\cite{zhang2023albert} & Recovery & Disaster Impact Assessment & Sentiment Classification & Encoder & No\\
\cite{varghese2024social} & Recovery & Disaster Impact Assessment & Sentiment Classification & Encoder & No\\
\cite{berbere2023exploring} & Recovery & Disaster Impact Assessment & Sentiment Classification & Encoder & No\\
\cite{li2025mining} & Recovery & Disaster Impact Assessment & Sentiment Classification & Decoder & No\\
\cite{white2024small} & Recovery & Recovery Plan Generation & Plan Generation & Decoder & No\\
\cite{lakhera2024leveraging} & Recovery & Recovery Plan Generation & Plan Generation & Decoder & No\\
\cite{contrerasassessing} & Recovery & Recovery Process Tracking & Sentiment Classification & Encoder & No\\

\end{longtable}

\footnotesize
\begin{longtable}{@{}>
{\raggedright\arraybackslash}p{2.5cm}p{1.5cm}p{1.5cm}p{1.6cm}p{1.8cm}p{1cm}p{1.5cm}p{1.5cm}@{}} 
\caption{Summary of publicly available datasets utilized in disaster management. For \textbf{Application}, "DI": Disaster Identification; "DInf": Disaster Information Coordination; "DIC": Disaster Issue Consultation; "DSA": Disaster Situation Assessment; "PAE": Public Awareness Enhancement; "DIA": Disaster Impact Assessment. For \textbf{Disaster Type}, "Mix" denotes the datasets contain various types of disasters. } \\
\vspace{-10mm}
\label{tab:dataset} \\
\toprule
\textbf{Dataset} & \textbf{Phase} & \textbf{Application} & \textbf{Task} & \textbf{Disaster Type} & \textbf{Modality} & \textbf{Used in}  & \textbf{\#Sample} \\ \midrule
\endfirsthead
\toprule
\textbf{Dataset} & \textbf{Phase} & \textbf{Application} & \textbf{Task} & \textbf{Disaster Type} & \textbf{Modality} & \textbf{Used in} & \textbf{\#Sample} \\ \midrule
\endhead
\bottomrule
\endfoot
CrisisLexT6~\cite{olteanu2014crisislex} & Response & DI & Classification & Mix & Text & \cite{mcdaniel2024zero} & 60,082 \\
CrisisLexT26~\cite{olteanu2015expect} & Response & DI, DInf & Classification & Mix & Text  & \cite{mcdaniel2024zero} & 27,933 \\
CrisisNLP~\cite{imran-etal-2016-twitter} & Response & DI, DInf & Classification & Mix & Text  & \cite{taghian2023disaster} & 52,656 \\
SWDM13~\cite{imran2013practical} & Response & DI, DInf & Classification & Mix & Text  & \cite{mcdaniel2024zero} & 1,543 \\
ISCRAM2013~\cite{imran2013practical} & Response & DI, DInf & Classification & Mix& Text & \cite{mcdaniel2024zero} & 3,617 \\
DRD~\cite{alam2021crisisbench} & Response & DI, DInf & Classification & Mix & Text   & \cite{mcdaniel2024zero} & 26,235 \\
DSM~\cite{alam2021crisisbench} & Response & DI & Classification & Mix & Text   & \cite{mcdaniel2024zero} & 10,876 \\
AIDR~\cite{imran2014aidr} & Response & DI, DInf & Classification & Mix & Text  & \cite{mcdaniel2024zero}& 7,411 \\
CrisisMMD~\cite{alam2018crisismmd} & Response & DI, DInf & Classification &Mix & Text, Image  & \cite{jain2024classification} & 16,058 \\
Multi-Crisis~\cite{sanchez2023cross} & Response & DI, DInf & Classification & Mix & Text & \cite{sanchez2023cross} & 164,625 \\
CrisisBench~\cite{alam2021crisisbench} & Response & DI, DInf & Classification & Mix & Text  &\cite{mcdaniel2024zero} & 109,796 \\
Eyewitness Messages~\cite{zahra2020automatic} & Response & DInf & Classification & Mix& Text & \cite{zahra2020automatic} & 14,000 \\
TREC Incident Streams~\cite{mccreadie2019trec} & Response & DI, DInf & Classification& Mix & Text & \cite{khattar2022camm} & 19,784 \\
HumAID~\cite{alam2021humaid} & Response & DInf & Classification & Mix & Text  & \cite{basit2023natural} & 77,000 \\
EPIC~\cite{stowe2018developing} & Response & DI, DInf & Classification & Mix& Text  & \cite{adesokan2023tweetace} & 3469 \\
Did You Feel It (DYFI)~\cite{mousavi2024gemini} & Response & DSA & Estimation & Earthquake & Text & \cite{mousavi2024gemini} & 750,000 \\
FloodDepth~\cite{akinboyewa2024automated} & Response & DSA & Estimation & Flood & Text, Image  & \cite{akinboyewa2024automated} & 150 \\
Behavioral Facilitation (BF)~\cite{yamamoto2022methods} & Response & DInf & Estimation & Earthquake & Text  & \cite{yamamoto2022methods} & 1,400 \\
\cite{fu2024extracting} & Preparedness & PAE & Extraction & Flood & Text& \cite{fu2024extracting} & 633 \\
\cite{ma2023ontology} & Preparedness & PAE & Extraction & Landslide & Text  & \cite{ma2023ontology} &  5,560 \\
DisasterMM~\cite{andreadis2022disastermm} & Response & DI, DInf & Classification, Extraction & Flood & Text  & \cite{mehmood2024named} &  6,672, 4,992 \\
\cite{suwaileh2022disaster}  & Response & DInf & Extraction & Mix & Text  & \cite{suwaileh2022disaster}  & 22,137 \\ 
Re´SoCIO~\cite{caillaut2024entity}  & Response & DInf & Extraction & Flood & Text  & \cite{caillaut2024entity} & 4,617 \\ 
~\cite{nguyen2022rationale}  & Response & DInf & Extraction & Mix & Text & \cite{nguyen2022rationale} & 32 \\ 
~\cite{vitiugin2022cross}  & Response & DInf & Generation & Mix & Text & \cite{vitiugin2022cross} & 5,791 \\
CrisisFACTS~\cite{mccreadie2023crisisfacts} & Response & DIC & Generation & Mix & Text & \cite{pereira2023crisis} & 748,466 \\
DisasterQA~\cite{rawat2024disasterqa} & Response & PAE, DIC & Generation & Mix & Text  & \cite{rawat2024disasterqa} & 707 \\
FFD-IQA~\cite{sun2023unleashing} & Response & DIC & Generation & Flood & Text, Image  & \cite{sun2023unleashing} & 22,422 \\
FloodNet~\cite{rahnemoonfar2021floodnet} & Recovery & DIA & Generation & Flood & Text, Image & \cite{sarkar2023sam} &  4,500 \\
\end{longtable}

\clearpage
\onecolumn

\end{document}